\documentclass[11pt,a4paper]{article}
\usepackage[hyperref]{acl2021}
\usepackage{times}
\usepackage{multirow}
\usepackage{latexsym}

\usepackage{microtype}

\aclfinalcopy 
\def\aclpaperid{59} 

\title{Generating clickbait spoilers with an ensemble of large language models}

\author{Mateusz Woźny$^1$ \and Mateusz Lango$^{1,2}$ \\
  $^1$ Poznan University of Technology,  Faculty of Computing and Telecommunications, Poznan, Poland  \\
  $^2$ Charles University, Faculty of Mathematics and Physics, Prague, Czech Republic\\
  \texttt{mateusz.wozny@student.put.edu.pl},   \texttt{mlango@cs.put.edu.pl} \\}

\date{}

\begin{document}
\maketitle
\begin{abstract}
Clickbait posts are a widespread problem in the webspace.
The generation of spoilers, i.e. short texts that neutralize clickbait by providing information that satisfies the curiosity induced by it, is one of the proposed solutions to the problem.
Current state-of-the-art methods are based on passage retrieval or question answering approaches and are limited to generating spoilers only in the form of a phrase or a passage.
In this work, we propose an ensemble of fine-tuned large language models for clickbait spoiler generation. Our approach is not limited to phrase or passage spoilers, but is also able to generate multipart spoilers that refer to several non-consecutive parts of text.
Experimental evaluation demonstrates that the proposed ensemble model outperforms the baselines in terms of BLEU, METEOR and BERTScore metrics.
\end{abstract}

\section{Introduction and related works}
Catchy headlines or social media posts designed to entice users to click, known as clickbait, are widespread on the internet. 
Although they often increase website traffic and generate revenue, they usually fall short of readers' expectations, wasting their time and causing disappointment~\cite {Molyneux}. 

To deal with this problem, \citet{rubin} proposed the clickbait detection task, which received some research attention~\cite{Potthast2016ClickbaitD,stop}.
More recently, \citet{hagen:2022} found that clickbaits can be neutralized by providing short texts that clarify what the reader can expect from the linked article, often making the clickbait uninteresting. 
They developed corpora of clickbait spoilers and classified them into three types: phrase spoilers (containing a single word or a short phrase), passage spoilers (a few sentences at most), and multi-part spoilers (containing many non-consecutive phrases and/or passages). See the examples in Tab.~\ref{tab:my_label}.

\citet{hagen:2022} also experimented with 20 approaches for clickbait spoiler generation, which were based on passage retrieval or extractive question answering algorithms. 
 However, all of these methods were only evaluated on phrase and passage spoilers, 
as they are not suitable for generating multi-part spoilers and their generation requires a specialised approach.

In this work, we demonstrate that all three types of spoilers can be effectively generated by means of conditional language generation with large language models.
We put forward a simple yet effective proposal of an ensemble of LLMs that selects the final spoiler by exploiting learning-to-rank techniques. 
Finally, we verify the performance of the proposed approach and investigate the possibility of combining it with previously developed methods that provide phrase and passage spoilers.

\begin{table}
    \centering
    \begin{tabular}{cp{27mm}p{26mm}}
     type & clickbait   & spoiler \\\hline
      phrase &You're missing this major way to save money 
   & promotional code\\
      passage & Scientists unearth big surprise near celebrated pyramids
& remains of a bustling port and barracks for sailors or troops
\\
multi & This is what REALLY happens when you don't brush your teeth &Bad breath, Coronary heart disease, Bleeding gums, (...)\\\hline
    \end{tabular}
    \caption{Abbreviated examples of spoilers and clickbaits from Webis-Clickbait-22 corpus~\cite{hagen:2022}}
    \label{tab:my_label}
\end{table}

\section{Ensemble of LLMs for clickbait spoiler generation}
The task of clickbait spoiler generation is defined as follows. 
For a given clickbait text $c$,  the content of the linked article $a$ and the requested spoiler type $t$, generate a textual spoiler $s$ whose goal is to make the clickbait $c$ uninteresting for the user by providing the additional information from the referred article $a$.
The possible types of spoilers $t$ are phrase, passage, and multi-part.

In this paper, we propose an ensemble of language models for clickbait spoiler generation.
The proposed approach consists of three steps: converting the text of a clickbait $c$ into a question $q$, generating candidate spoilers from various prompted large language models, and finally selecting the final spoiler by a trained scoring model.

\subsection{Converting clickbaits to questions}
\newcommand{\n}{\textbackslash n }
\label{sec:convertq}
Clickbaits usually take the form of declarative or exclamatory sentences.
In contrast, question answering, which is one of the most related tasks according to related works, naturally deals with problems structured as interrogative sentences.
Due to their popularity, QA datasets are often used as a part of LLMs' (pre)training sets,  enabling better knowledge transfer for these tasks.
Therefore, to better exploit knowledge acquired by LLMs during pretraining, we convert each clickbait into a question before passing it for further processing.

The conversion is made in a zero-shot fashion using the recent Vicuna language model~\cite{vicuna2023}. 
For each clickbait $c$, we construct the following prompt: "Below is a sentence from which write a question.\textbackslash n         Sentence: $c$ \textbackslash n Question:", where \textbackslash n is the sign of a new line.
The resulting question $q$ is generated by initializing the language model with the prompt and completing the text with the greedy search algorithm until the sign of a new line is generated.

\subsection{Generating spoilers with LLMs}
\label{sec:generating}
The next step of our approach is to use a set of different pretrained language models to produce a diversified set of spoiler candidates. 

Each component of our ensemble is fine-tuned on the standard language modeling task using an adapter-based approach LoRA~\cite{hu2021lora}. Such transfer learning approaches allow parameter-efficient fine-tuning by leaving all the pretrained weights unchanged and modifying  the model operations by adding shallow, trainable feed-forward networks between the transformer layers. The results of these feed-forward networks are incorporated into the transformer architecture by adding their output to the output of successive transformer layers.
Such fine-tuning approaches have proven to be well-suited for relatively small supervised datasets like ours~\cite{HoulsbyGJMLGAG19}.

In order to create training corpora for the language modeling task, each training example was converted to a textual form by filling in the hand-designed prompt template: "Below is a question paired with a context for which you should generate an answer. Write an answer with type $t$ that appropriately completes the question.\n Question: $q$ \n Context: $a$ \n Answer: $s$\n".
During training, cross-entropy loss was optimized, i.e. $$-  \sum_{i=1}^n \log P(w_i|w_1, w_2, ..., w_{i-1})$$
where $w_i$ is the $i$-th token of the filled template and $n$ is its length.
During testing, standard greedy decoding was used to retrieve the clickbait spoiler.    

\subsection{Selecting the final spoiler with scoring model}
\label{sec:ranking}
After generating several clickbait spoiler candidates, the final step of the approach is to select the most appropriate spoiler by running a trainable model that evaluates them.
This problem can be viewed as a learning-to-rank problem~\cite{ltr}, where our goal is to construct the ranking of spoiler candidates and later select the best candidate i.e. the spoiler at the top of the ranking.
We experiment with two popular learning-to-rank approaches:  1) a pointwise approach, which assigns a score to each candidate and later uses it to sort the list of candidates. 2) a pairwise approach, which compares all pairs of candidates and decides which spoiler from the pair is more suitable.

\subsubsection{Pointwise approach}
To evaluate each candidate, we develop a regressor that tries to predict the value of BLEU score for each spoiler.
As a regressor fine-tuned DeBERTa model~\cite{he2021deberta} with one linear layer on top of CLS token is used.
The input to the model consists of a question (clickbait) $q$, candidate spoiler $s$, and article $a$, separated by the sign of a new line and concatenated into one input text.
The output of the model is the predicted BLEU score.

The training data for the regressor was generated by running all the LLMs used in the ensemble on the training data and evaluating their BLEU score against the available gold standard.
During training, the classical sum of squared errors was optimized.

\subsubsection{Pairwise approach}
The second method considered for selecting the best spoiler among the candidates is the pairwise approach.
This approach relies on a classifier that, for a given pair of spoiler candidates, decides which of them is more suitable.
More specifically, the BERT-based classifier receives the same input as in the pointwise approach, but with two spoiler candidates $s_1$ and $s_2$.
The output of the binary classification model is 1 if $s_1$ is better than $s_2$ in terms of BLEU, and 0 otherwise.

The classifier is trained on the generated data as follows.
First, all LLMs used in the ensemble were run on the training set, generating a collection of spoiler candidates for each clickbait.
Later, all possible pairs from each collection were considered and converted into binary classification instances by comparing the BLEU scores of the candidates. 
The pairs containing spoilers with BLEU$=0$ or pairs containing identical spoilers were removed from the training data.
During training, the standard cross-entropy loss was optimised.

\section{Experiments}
We have performed computational experiments aimed at verifying the effectiveness of ensembling LLMs with pointwise and pairwise rankers, and comparing its effectiveness with the previous SOTA methods based on question answering.
In addition, we also investigate the possibility of combining the previous QA approaches, which are best suited for passage and phrase spoiler types, with the proposed approach for multi-part spoilers.

\subsection{Experimental setup}
We experiment with an ensemble of three fine-tuned LLMs, which were constructed from two pretrained models: LLaMA~\cite{llama} and Vicuna~\cite{vicuna2023}. Both of these models are open-source and were fine-tuned using the prompt described in Sec.~\ref{sec:generating}.
However, we observed additional improvements with Vicuna model while using specially tailored prompts for each spoiler type separately (see appendix), therefore we also report the results of this fine-tuned model and use it as a part of the ensemble.

The obtained results were compared against the performance of two extractive QA approaches, which on top of pretrained encoder perform begin/end span classification\footnote{This is the default fine-tuning approach for QA-task of BERT~\cite{devlin-etal-2019-bert}, more details therein.}. 
These approaches are based on RoBERTa~\cite{roberta} and DeBERTa~\cite{he2021deberta} models since among 20 approaches compared on clickbait spoiler generation task by \citet{hagen:2022} these two were the most effective ones.
Note, that the results reported for these approaches in this work are significantly lower than therein, since we report the performance over all three types of spoilers, including multipart. 

The ensemble approach with the pointwise ranker used DeBERTa-based regressor, which obtained MSE of 0.384 on test set. Similarly, the classifier  used in the pairwise approach achieved balanced accuracy of 90,8\%  on the test data.

Following earlier works, we evaluated the approaches with three metrics: BLEU~\cite{bleu}, METEOR~\cite{meteor}, and BERT Score~\cite{bertscore}.
All experiments were performed on a single Nvidia A100 GPU.

Some additional experiment details and results can be found in the online appendix\footnote{\url{https://www.cs.put.poznan.pl/mlango/publications/inlg23.pdf}}.

\subsection{Comparing the proposed approaches with related works}
\begin{table*}[t]
\centering
\begin{tabular}{llllll}
\hline
                        &       &        & \multicolumn{3}{c}{BERT Score} \\
Model                   & BLEU  & METEOR & Precision   & Recall  & F1     \\
\hline
RoBERTa                & 31,78 & 0,387  & 0,904       & 0,883   & 0,893  \\
DeBERTa                & 32,20 & 0,398  & 0,907       & 0,884   & 0,894  \\
LLaMA 13B               & 37,70 & 0,474  & 0,895       & 0,901   & 0,897  \\
Vicuna 13B              & 38,80 & 0,481  & 0,898       & 0,903   & 0,900  \\
Vicuna 13B with type-based prompts             & 40,02 & 0,492  & 0,899       & 0,905   & 0,901  \\
Ensemble with pairwise ranker    & 40,76 & 0,500  & 0,901       & 0,907   & 0,904  \\
Ensemble with pointwise ranker & 42,13 & 0,517  & 0,902       & 0,909   & 0,905 \\\hline
\end{tabular}
\caption{\label{main-res}
The experimental results of previous state-of-the-art QA-based methods compared with our ensembling approaches and LLMs. All the metrics are computed on Webis Clickbait 22 corpora~\cite{zenodo}.
}
\end{table*}
\begin{table*}[t]
\centering
\begin{tabular}{llllll}
\hline
                        &       &        & \multicolumn{3}{c}{BERT Score} \\
Model                   & BLEU  & METEOR & Precision   & Recall  & F1     \\\hline
DeBERTa                 & 32,20 &  0,398  & 0,907       & 0,884   & 0,894  \\
DeBERTa trained on questions    & 37,82 & 0,451  & 0,913       & 0,895   & 0,903  \\
Baseline ensemble       & 42,28 & 0,506  & 0,910       & 0,908   & 0,909  \\
Ensemble with pairwise ranker    & 43,57 & 0,520  & 0,912       & 0,911   & 0,911  \\
Ensemble with pointwise ranker & 44,45 & 0,532  & 0,911       & 0,913   & 0,911 \\\hline
\end{tabular}
\caption{\label{merge-res}
The experimental results of the ensemble that combines previously proposed methods for passage and phrase spoiler types (DeBERTa) with the newly proposed approaches for clickbait spoiler generation. 
}
\end{table*}
The results of QA-based approaches, ensemble models as well as individual fine-tuned LLMs can be found in Table~\ref{main-res}.
The best-performing approach according to BLEU, METEOR, and BERT Score F1 is the proposed ensemble with a pointwise ranker.
This ensemble provides the improvement of approx.~2 percentage points in terms of BLEU and METEOR over the best of its components i.e.~Vicuna model with  specific  prompts for each spoiler type.
The second-best approach was the ensemble with pairwise ranker which offered limited improvement over the individual LLMs.

Overall, each of the approaches using LLMs obtained better results than previous state-of-the-art approaches based on extractive question answering. 
The only metric on which the QA-based approaches (RoBERTa and DeBERTa) stand out is   BERTScore Precision. Still, BERTScore Recall is higher for LLM and ensemble approaches, making them more effective in terms of F1 measure, which combines both precision and recall.

\subsection{Combining  previous SOTA models with the proposed ones}

In the final experiment, we decided to verify whether it is possible to obtain even better results by combining our approaches with QA-based models previously designed for clickbait spoilers of phrase and passage types only.

As a QA model, we use the fine-tuned DeBERTa model, since it gave the best results both in our experiments from the previous section and in the experiments of~\citet{hagen:2022}.
As we mentioned in Sec.~\ref{sec:convertq}, our approaches generate spoilers for clickbaits converted into interrogative sentences in order to facilitate better knowledge transfer from the pre-trained models.
We also fine-tuned DeBERTa on such preprocessed data and found improved performance.
Therefore, this model (later denoted DeBERTa-q or "DeBERTa trained on questions") was used for ensembling.

The operation of ensembles reported in this section slightly differs from what was described in Sec~\ref{sec:ranking} to better account for QA-based approaches' suitability for passage and phrase spoiler types.
If the generated spoiler is of mutli-part type, the list of candidate spoilers is generated as previously, i.e. from three fine-tuned LLMs.
However, if the generated spoiler is of phrase/passage type then only spoiler candidates from DeBERTa-q and Vicuna with customised prompts are considered.
As previously, the selection of the best spoiler among the candidates is performed by a ranker.

The results of these approaches are reported in Table~\ref{merge-res}.
It can be seen that DeBERTa-q achieves significantly better results than DeBERTa for all metrics considered. However, the ensemble with LLMs provides further significant improvements.

As a form of sanity check for our ensemble approach that uses a ranker to select the best spoiler, we have also implemented a trivial ensemble (baseline ensemble) that uses DeBERTa-q for all passage and phrase spoilers and Vicuna with type-based prompts for multi-part spoilers.
Although such a form of ensembling is also advantageous in that the performance obtained is better than that of the individual models, using any variant of the proposed ensemble with a ranker still improves the results. 
For example, for the METEOR measure, the improvement is almost 2\% for the pairwise approach and 3\% for the pointwise approach.

\begin{table*}[t]
\begin{tabular}{llrrrrr}
\hline
                                                &         &       &        & \multicolumn{3}{c}{BERT Score} \\
Model                                           & spoiler type    & BLEU  & METEOR & Recall   & F1     & Precision  \\\hline
\multirow{3}{*}{DeBERTa}                        & phrase  & 56,00 & 0,569  & 0,934    & 0,932  & 0,931      \\
                                                & passage & 20,10 & 0,304  & 0,858    & 0,869  & 0,883      \\
                                                & multi-part   & 2,07  & 0,204  & 0,822    & 0,860  & 0,902      \\\hline
\multirow{3}{*}{DeBERTa trained on questions}   & phrase  & 62,50 & 0,627  & 0,946    & 0,942  & 0,939      \\
                                                & passage & 21,60 & 0,319  & 0,861    & 0,872  & 0,885      \\
                                                & multi-part   & 4,10  & 0,224  & 0,829    & 0,862  & 0,900      \\\hline
\multirow{3}{*}{Ensemble with pairwise ranker}  & phrase  & 63,30 & 0,617  & 0,945    & 0,943  & 0,942      \\
                                                & passage & 30,00 & 0,416  & 0,882    & 0,887  & 0,894      \\
                                                & multi-part   & 27,30 & 0,525  & 0,894    & 0,886  & 0,879      \\\hline
\multirow{3}{*}{Ensemble with pointwise ranker} & phrase  & 65,30 & 0,645  & 0,950    & 0,946  & 0,942      \\
                                                & passage & 30,30 & 0,424  & 0,884    & 0,887  & 0,892      \\
                                                & multi-part   & 28,60 & 0,522  & 0,894    & 0,887  & 0,881     \\\hline
\end{tabular}
\caption{The performance of clickbait spoiler generation models for each spoiler type. }
\label{tab:results}
\end{table*}

\begin{table*}[t]
\begin{tabular}{p{3cm}p{4.3cm}p{3cm}p{4.3cm}}
\hline
Clickbait                                     & Reference                                                                                                                                                        & DeBERTa                       & Vicuna                                                                                                                                                                          \\\hline
Agency might plant a garden on the moon.      & NASA                                                                                                                                                                           & \textit{NASA}                         & lunar sunlight                                                                                                                                                                 \\
{10 habits of incredibly happy people} & 1 they slow down to appreciate life's little pleasure 2 they exercise 3 they spend money on other people (...) & they have a growth mindset & \textit{1 they slow down to appreciate life's little pleasure 2 they exercise 3 they spend money on other people (...)}
\\\hline
\end{tabular}
\caption{Two examples of spoilers generated by different methods. The spoilers in italics were selected by pointwise ranker (The ensemble also includes the LLaMA model, the output of which is not shown due to page limits.).}
\label{tab:examples}
\end{table*}

Table~\ref{tab:results} presents a more detailed analysis of the performance of the spoiler generation methods, i.e.~the results are reported separately for each spoiler type. Although the newly proposed ensemble approaches achieve better performance for each spoiler type, the most significant improvement is observed for the most difficult multi-part spoilers, which are the focus of our paper. For example, the best ensemble model achieves an improvement of over 25 in terms of BLEU score over the previous DeBERTa model.

Two examples of generated spoilers are presented in Table~\ref{tab:examples} (more examples in the appendix). It can be observed that the previous SOTA approach (DeBERTa) fails to extract the correct spoiler of the multi-part type, but Vicuna model generates it correctly. In contrast, DeBERTa extracts the correct phrase spoiler and Vicuna fails to do so. In both cases, the pointwise ranker was able to indicate the correct spoiler.

\section{Summary}
In this paper, we have shown  that using fine-tuned LLMs can be a simple, yet effective way of dealing with clickbait spoiler generation for all three spoiler types considered, i.e.~including multi-part spoilers.
We also demonstrated that ensembling several such models with a ranker that selects the most suitable spoiler leads to improved results, especially when using a pointwise ranker.
Finally, we show that combining state-of-the-art approaches for phrase and passage spoiler types based on question answering with the newly proposed ones based on LLMs leads to further improvements. 

\paragraph*{Supplementary Materials Availability Statement:} 
Source code is available on Github repository\footnote{\url{https://github.com/mateusz-wozny/spoiler_generation}}.
All experiments were performed on Webis Clickbait 22 corpora which is available on Zenodo~\cite{zenodo}.

\paragraph*{Acknowledgments} M. Woźny is supported by AI Tech project sponsored from POPC programme POPC.03.02.00-00-0001/20. M. Lango is supported by the European Union (ERC, NG-NLG, 101039303).
\bibliographystyle{acl_natbib}
\bibliography{anthology,acl2021}


\end{document}